\documentclass[runningheads]{llncs}
\usepackage{graphicx,amsmath,amssymb, cancel, bm}

\begin{document}

\newcommand{\bx}{\mathbf{x}}
\newcommand{\by}{\mathbf{y}}
\newcommand{\bbf}{\mathbf{f}}
\newcommand{\bu}{\mathbf{u}}
\newcommand{\bE}{\mathbb{E}}
\newcommand{\bR}{\mathcal{R}}
\newcommand{\bN}{\mathcal{N}}
\newcommand\blfootnote[1]{%
  \begingroup
  \renewcommand\thefootnote{}\footnote{#1}%
  \addtocounter{footnote}{-1}%
  \endgroup
}
\title{Deep Gaussian Processes with Convolutional Kernels}

\author{Vinayak Kumar$^*$\inst{1}\and
Vaibhav Singh$^*$\inst{1}\and
P.~K.~Srijith\inst{1} \and
Andreas Damianou$\dagger$\inst{2}}

\authorrunning{V. Kumar et al.}

\institute{Department of Computer Science and Engineering, Indian Institute of Technology, Hyderabad, India \and
Amazon Research, Cambridge, United Kingdom \\
\email{\{vinayakk,cs16mtech11017,srijith\}@iith.ac.in}, \email{damianou@amazon.com}}

\maketitle              

\begin{abstract}
\blfootnote{$^{*}$Equal Contribution.}
\blfootnote{$^{\dagger}$A. Damianou contributed to this work prior to joining Amazon.}

Deep Gaussian processes (DGPs) provide a Bayesian non-parametric alternative to standard parametric deep learning models. A DGP is formed by stacking multiple GPs resulting in a well-regularized composition of functions. The Bayesian framework that equips the model with attractive properties, such as implicit capacity control and predictive uncertainty, makes it at the same time challenging to combine with a convolutional structure. This has hindered the application of DGPs in computer vision tasks, an area where deep parametric models (i.e.\ CNNs) have made breakthroughs. Standard kernels used in DGPs such as radial basis functions (RBFs) are insufficient for handling pixel variability in raw images. In this paper, we build on the recent convolutional GP to develop Convolutional DGP (CDGP) models which  effectively capture image level features through the use of convolution kernels, therefore opening up the way for applying DGPs to computer vision tasks. Our model learns local spatial influence and outperforms strong GP based baselines on multi-class image classification. We also consider various constructions of convolution kernel over the image patches, analyze the computational trade-offs and provide an efficient framework for convolutional DGP models. The experimental results on image data such as MNIST, rectangles-image, CIFAR10 and Caltech101 demonstrate the effectiveness of the proposed approaches.

\end{abstract}

\section{Introduction}

Deep learning models have made tremendous progress in computer vision problems through their ability to learn complex functions and representations~\cite{dl16}. They learn complex functions mapping some input x to output y through composition of linear and non-linear functions. However, popular deep learning models based on convolutional and recurrent neural networks have significant limitations.
The parametric form of the functions lead  them to have millions of parameters to estimate which is less suitable for problems where the data are scarce. Deep learning models though probabilistic in nature, do not provide any uncertainty estimates on its predictions. Knowledge of uncertainty helps in better decision making and is crucial in high risk applications such as disease diagnosis and autonomous driving~\cite{gal16}. Another major limitation with the existing deep learning networks is model selection. Developing an appropriate deep learning model to solve a problem is time consuming and computationally expensive. Deep Gaussian processes (DGPs)~\cite{damianou13} constitute a deep Bayesian non-parametric approach based on Gaussian processes (GPs) and have the potential to overcome the aforementioned limitations.

The original DGP model was introduced by \cite{damianou13,damianou:thesis15} inspired by the hierarchical GP-LVM structure \cite{lawrence2007hierarchical} and variations have emerged in recent years, mainly differing in the employed inference procedure. While \cite{damianou13} employs a mean field variational posterior over the latent layers, \cite{Dai:VAEDGP16} extends this formulation with amortized inference, \cite{hensman2014nested} considers a nested variational inference approach, \cite{bui2016deep} uses an approximate Expectation Propagation procedure. Further, \cite{randombonila} achieves scalability through random Fourier features while the approach of \cite{dsvi} considers the variational posterior to be conditioned over the previous layer, preserving  correlations across the layers, and uses a doubly stochastic variational inference approach.

All the DGP models use kernels such as  radial basis function (RBF) which is inadequate for problems in computer vision, such as object detection. They fail to capture  wide variability of objects in images due to pose, illumination and complex backgrounds. RBF captures similarity between images on a global scale and is not invariant to unwanted variations in the image. On the other hand, convolutional neural networks (CNN)~\cite{dl16} learn image representations from raw pixel data which are invariant to such perturbations in the image. They learn features important for the object detection task by successively convolving the representations by filters, applying non-linearity and performing feature pooling. We propose to use convolutional kernels~\cite{haussler99} in DGPs to learn salient features from the images which are invariant to transformations. This is different from recent works which combine CNNs and GPs in hybrid mode, such as~\cite{wilson16,calibDCGP,bradshaw2017adversarial}. In particular,~\cite{calibDCGP} replaces the fully connected layers of a CNN with GPs, aiming at obtaining well-calibrated probabilities. While, in deep kernel learning~\cite{wilson16}, the kernel in GPs are computed using deep neural networks. In contrast, our approach brings the convolutional structure inside the deep GP model, through kernels, and remains fully non-parametric.

Convolutional kernels could effectively learn rich representations of the data. The similarity between structured objects such as images are computed by considering the similarity of the sub-structures in the object which makes them invariant to transformations in the image. They have been used to compute similarities between structured objects such as graphs and trees~\cite{vish10,collins01}. Recently, they were used as a covariance function in GPs and were found to be very effective  for object recognition tasks~\cite{convgp}. Here, the kernel computations between images are done  by summing the base kernel acting over different patches of the images. 

We introduce convolutional kernels in the DGP framework in order to extract discriminative features from images for object classification. Our work builds on the convolutional GP~\cite{convgp} and extends it for the deep learning case, allowing the resulting model to additionally perform hierarchical feature learning. We consider various DGP architectures obtained by stacking together convolutional and RBF kernels in various combinations. Further, we consider variants of the convolutional kernel such as weighted convolutional kernels which provide more discriminative features, and  combination of RBF kernels as the base kernel. 
Convolutional kernels are computationally expensive as they require performing summation over all  patches of the image. We propose an approach to improve the computational efficiency by random sub-sampling of the patches. We demonstrate the effectiveness of the proposed approaches for image classification on benchmark data sets such as MNIST, Rectangles-image, CIFAR10 and Caltech101. The experiments show that DGP models typically achieve better generalization performance by using convolutional kernels compared to state-of-the-art shallow GP models.

\section{Background}
We consider the image classification problem with $C$ classes and  $N$ training data points, $X = \{\bf{x_i}\}_{i=1}^N$ and the corresponding labels $\bf{y} = \{y_i\}_{i=1}^N$, where $\bf{x_i} \in \mathcal{R}^{W \times H}$ and $y_i \in \mathcal{Y} = \{1,2, \ldots C\}$. Assume there exists  a latent function $f : \mathcal{R}^{W \times H} \rightarrow \mathcal{Y}$ mapping the training data to outputs. In a Bayesian setting, we strive to learn a posterior distribution over this function, so that we can use it to compute the predictive distribution over the test labels. It helps one to make sound predictions about the test data labels, taking into account the uncertainty about them. Gaussian processes provide a Bayesian non-parametric approach to perform classification. In this section, we summarize Gaussian process classification and Deep Gaussian Processes (DGP) that will lay the groundwork for our model.  

\subsection{Gaussian Process}
\label{sect:gp}

A GP is defined as a collection of random variables such that any finite subset of which is Gaussian distributed ~\cite{GPMbook}. It allows one to specify a prior distribution over real valued functions $f$, represented as
$f(\bx) \sim \mathcal{GP}(m(\bx),k(\bx,\bx'))$
where $m(\bx)$ is the mean function and $k(\bx,\bx')$ provides the covariance across the function values at two data points $\bx$ and $\bx'$.

The kernel function determines various properties of the function such as  stationarity, smoothness etc. A popular kernel function is the radial basis function (RBF) (squared exponential kernel), as it can model any smooth function. It is  given by $\sigma_{f}^{2}\exp(-\frac{1}{2 \kappa}||\bx - \bx'||^2)$ where the  length scale $\kappa$ determines the variations in function values across the inputs.

For multi-class classification problems, we associate a separate function $f_c$ with each class $c$. An independent GP prior is placed over each of these functions,  $f_c(\bx) \sim {\mathcal{GP}}(m_c(\bx),k(\bx,\bx'))$.
Let ${\bf f_c}=[f_c(\bx_1),f_c(\bx_2),\cdots,f_c(\bx_N)]$ 
be a column vector indicating function values at the input data points for a class $c$.  
Further, let $F$ be the matrix formed by  stacking all column vectors $\{\bf f_c\}_{c=1}^C$ , with $F_{n,c}$ representing the latent function value of $n^{th}$ sample belonging to class $c$ and $F_n$ representing the vector of latent function values over classes for the $n^{th}$ sample. 
The GP prior over $F$ takes the following form :  $p(F) = \prod\limits_{c=1}^C \mathcal{N} (\bbf_c; m_c(X), K_{XX})$, where $K_{XX}$ is the $N \times N$ covariance matrix formed by evaluating kernel over all pairs of training data points. 
For a data point $n$, the likelihood of it belonging to class $c$, $p(y_n = c | F_n)$, is obtained by considering  a soft-max link function.
The posterior distribution over $F$ is obtained by combining the prior and the likelihood using Bayes theorem:
\begin{equation*}
p(F|{\bf y}) = \frac{\prod_{n=1}^N p(y_n | F_n) p(F)}{p(\bf{y})} .
\end{equation*}

In GP multi-class classification, the posterior distribution cannot be computed in closed form due to the non-conjugacy between likelihood and prior.  Learning in GPs involves learning the kernel hyper-parameters by maximizing the evidence $p(\by) = \int \prod_{n=1}^N p(y_n | F_n) p(F) dF$, which also cannot be computed in closed form. 
The posterior distribution can be approximated as a Gaussian using approximate inference techniques such as Laplace approximation~\cite{gpc1} and variational inference~\cite{gpc2,hensman2015scalable,srijith16}. The Gaussian approximated posterior is then used to make predictions on the test data points. Variational inference has received a lot interest recently as it does not suffer from convergence problems unlike Markov chain Monte Carlo techniques and it provides a posterior approximation quickly by solving an optimization problem. It is scalable to large data sets and amenable to distributed processing. It also provides a lower bound on the marginal likelihood which can be used to perform model selection. The variational inference approach learns an approximate posterior distribution $q(F)$ by minimizing the KL divergence between $q(F)$ and $p(F|{\bf y})$. Choosing a mean field family of variational distributions, $q(F)$ factorizes across dimensions(or columns), i.e $q(F) = \prod q({\bbf_c})$. Each variational factor $q(\bbf_c)$ is assumed to be a Gaussian with variational parameters,  mean vector $\bf \mu_c$ and covariance $\Sigma_c$. In the variational inference framework, minimizing the KL divergence with respect to the variational parameters is equivalent to maximizing the so-called variational  \textbf{E}vidence \textbf{L}ower \textbf{BO}und\textbf{(ELBO)} which is given by
\begin{equation}
\label{velbo}
L( \{{\bf \mu_c} , \Sigma_c\}_{c=1}^C) =  \bE_{q(F)}[\text{log} \prod_{n=1}^N p(y_n |  F_n )] - \sum\limits_{c=1}^C \text{KL} (q(\bbf_c)\parallel p(\bbf_c)). 
\end{equation}
The variational parameters $\{{\bf \mu_c},\Sigma_c\}_{c=1}^C$ and the kernel hyperparameters $\{\sigma_{f}^2,l\}$ are learnt by jointly maximizing the variational lower bound  in eq. \eqref{velbo} using any gradient based approach.  

The KL divergence term in eq. \eqref{velbo} involves inversion of the covariance matrix $K_{XX}$ which scales as ${\mathcal{O}}(N^3)$ computationally. Therefore, we opt for the variational sparse Gaussian process approximation~\cite{GPbigdata,titsias2009variational} which reduces the computational complexity to ${\mathcal{O}}(NM^2)$ , where $M \ll N$ represents the number of inducing points.   %
Specifically, the variational sparse approximation expands the latent function space with $M$  inducing variables $\mathbf{u} \in \bR^M$ which are latent function values at inducing points $Z = \{\mathbf{z}_i\}_{i=1}^M$. Within the context of GP multi-class classification, we additionally have the inducing variable outputs $\bu_c$ for each class $c$ which are stacked together to form the matrix $U \in \bR^{M  \times C}$. 
The joint GP prior over $\{f,u\}$ is then
\begin{equation}
\begin{bmatrix}
\bbf_c \\
\bu_c \\
\end{bmatrix}
\sim 
\bN(
\begin{bmatrix}
\bbf_c \\
\bu_c \\
\end{bmatrix}
;
\begin{bmatrix}
 m_c(X)\\
 m_c(Z)\\
\end{bmatrix}
,
\begin{bmatrix}
K_{XX}&K_{XZ}\\
K_{XZ}^\top &K_{ZZ}
\end{bmatrix}
),
\end{equation}
where $K_{XZ}$ is the $N \times M$ covariance matrix over training inputs $X$ and inducing inputs $Z$ and  $K_{ZZ}$ is the $M \times M$ covariance matrix over inducing points $Z$. The conditional distribution of $\bbf_c$ given $\bu_c$ is given by
\begin{equation*}
p(\bbf_c|\bu_c, X, Z)=\bN(\bbf_c; m_c(X) + K_{XZ}K_{ZZ}^{-1}(\bu_c - m_c(Z)), K_{XX}-K_{XZ}K_{ZZ}^{-1}K_{XZ}^\top)
\end{equation*}
and the marginal distribution over $\bu_c$ is $p(\bu_c) = \bN(\bu_c;m_c(Z), K_{ZZ})$. The variational sparse approximation of ~\cite{titsias2009variational} considers a joint variational posterior over $\{\bbf_c,\bu_c\}$ in factorized form and is written as  $q(\bbf_c,\bu_c)=p(\bbf_c|\bu_c,X)q(\bu_c)$.
Assuming Gaussian variational factors for inducing points $q(\bu_c) = \bN(\bu_c; {\bf m_c} ,  S_c)$, the variational lower bound (ELBO)  can be derived as 
\begin{equation}
L( \{{\bf m_c} , S_c\}_{c=1}^C) = \bE_{q(F)}[\log \prod_{i=1}^N p(\by_n|F_n)]- \sum\limits_{c=1}^{C}\text{KL}(p(\bu_c) || q(\bu_c)).
\end{equation}
Following ~\cite{hensman2015scalable}, the variational posterior $q(F) = \prod_{c=1}^C q(\bbf_c)$ and $q(\bbf_c)$ is obtained by integrating out $\bu_c$ from $p(\bbf_c|\bu_c)q(\bu_c)$ and is given by $\bN(\bbf_c; {\bf \tilde{m}_c}, \tilde{V}_c)$ , where ${ \bf \tilde{m}_c} =\it m(X) + K_{XZ}K_{ZZ}^{-1}(\bf m_c - \it m(Z)) \text{ and } \tilde{V}_c = K_{XX}- k_{XZ}K_{ZZ}^{-1}(K_{ZZ}- S_c ) K_{ZZ}^{-1}K_{ZX}$. 
The Expected Log likelihood term above is intractable due to non-conjugate likelihood (softmax in this case). One could apply a quadrature~\cite{hensman2015scalable} or reparameterization-based~\cite{autoencodingVB} monte carlo sampling scheme to approximate this.

\subsection{Deep Gaussian Process}
\label{sect:dgp}

Deep Gaussian processes (DGPs)~\cite{damianou13,damianou:thesis15,dsvi} learn complex functions by stacking GPs one over the other resulting in a deep architecture of GPs. The function mapping one hidden layer to the next in DGPs is more expressive and data dependent compared to the pre-fixed sigmoid non-linear function used in standard parametric deep learning approaches. In addition, it is devoid of large number of parameters but only a few kernel hyper-parameters and few variational parameters (few due to sparse GP approach). Deep GPs do not typically overfit on small data due to Bayesian model averaging, and the stochasticity inherent in GPs naturally allows them to handle uncertainty in the data. Furthermore, by using a specific kernel which enables automatic relevance determination, one can automatically learn the dimensionality of hidden layers (number of neurons) \cite{damianou:thesis15}. This overcomes the model selection problem in deep learning to a great extent. 

DGPs consider the function mapping input to output to be  represented as a composition of functions, $f(\bx) = f^L \circ (f^{L-1} \ldots \circ (f^1 (\bx))) $, assuming there are $L$ layers. The $l^{th}$ layer consists of $D^l$ functions $f^l = \{ f^l_j\}_{j=1}^{D^l}$ mapping representations in layer $l-1$ to obtain $D^l$ representation for layer $l$. Independent GP priors are placed over the function $f^l_j$ producing $j^{th}$ representation in layer $l$, $f_{j}^{l}(\cdot)  \sim {\mathcal{GP}}(\it m_{j}^{l}(\cdot) , k^{l}(\cdot,\cdot))$. The $j^{th}$  function in layer $l$,  $f^1_j$, acts on the input data point $\bx_i$ to produce the representation $F^1_{i,j} = f^1_j(\bx_i)$. In general, the $j^{th}$ function in layer $l$, $f^l_j(\cdot)$ acts on the representation of the data point $\bx_i$ at layer $l-1$, $F^{l-1}_{i}$ to produce the  representation $F^l_{i,j} = f^l_j(F^{l-1}_{i})$. Let $\bbf^l_j$ denote the $j^{th}$ representation at layer $l$ computed over all inputs. The final layer $L$ will have $C$ functions corresponding to the classes and these functions values are squashed through a soft-max function to produce the class probabilities.

We follow the DGP variant presented in \cite{dsvi} where the noise between layers is absorbed into the kernel. The kernel function associated with a GP in layer $l$ is defined as $k^l(F^l_{i},F^l_{j})= {\sigma^l_{f}}^2 \exp(\frac{-1}{2\kappa^l} ||F^l_{i} - F^l_{j}||^2 ) + {\sigma^l_n}^2\delta_{ij} $. 
Following the variational sparse Gaussian process approximation  as explained in the section \ref{sect:gp}, each layer $l$ is associated with inducing variables $\{U^l\}$ which are  function values over $M$ inducing points $Z^l$ associated with layer $l$, $Z^l = \{{\bf z}^l_i \}_{i=1}^M$. Let $\bu^l_j$ represent the inducing variables associated with the $j^{th}$ representation at layer $l$.
The number of inducing points are kept fixed for all layers (only for convenience) as $M$ and a joint GP prior is considered over latent function values and inducing points.   
The joint distribution $p(\by,F,U)$ is given by
\begin{equation}
\underbrace{\hbox{$\prod\limits_{n=1}^{N}P(y_n|F^{L}_{n})$}}_{\hbox{Likelihood}}
\underbrace{\hbox{$\prod\limits_{l=1}^{L} \prod\limits_{j=1}^{D^l}p(\bbf_j^l|\bu_j^l, F^{l-1}, Z^l)p(\bu_j^l|Z^l)$}}_{\hbox{Deep GP Prior}},
\end{equation}
where a deep GP prior is put recursively over the entire latent space with $F^0 = X$ and a soft-max likelihood is used for classification. The conditional above is:
\begin{eqnarray}
\label{eqn:dgp_cond_prior}
& & p(\bbf_j^l|\bu_j^l,F^{l-1}, Z^l)=\bN(\bbf_j^l; mean(\bbf_j^l), cov(\bbf_j^l) ) \quad \text{where} \\
& &mean(\bbf_j^l) = m_j^l(F^{l-1}) + K^l_{F^{l-1}Z^l}(K^l_{Z^l Z^l})^{-1}(\bu_j^l - m_j^l(Z^l))\nonumber\\
& & cov(\bbf_j^l) = K^l_{F^{l-1}F^{l-1}}-K^l_{F^{l-1}Z^l}(K^l_{Z^lZ^l})^{-1}(K^l_{F^{l-1}Z^l})^\top\nonumber
\end{eqnarray}

The posterior distribution $p(F,U|\by)$ and marginal likelihood $p(\by)$ cannot be computed in closed form   
due to the intractability in obtaining the marginal prior over $\{F^l\}_{l=2}^L$. This involves integrating out the previous layer, which is present in a non linear manner inside the covariance matrices ($K^l_{F^{l-1}F^{l-1}}$) appearing in (\ref{eqn:dgp_cond_prior}). Along with non-conjugate likelihood, this brings in additional difficulty to the DGP model. Multiple approaches have been suggested in the literature for achieving tractability in DGPs, such as variational inference~\cite{damianou13,hensman2014nested,dsvi}, amortized inference~\cite{Dai:VAEDGP16}, expectation propagation~\cite{bui2016deep} and random Fourier features~\cite{randombonila}. Here we follow the variational inference approach, and we assume the variational posterior to be having form $q(F,U)=\prod\limits_{l=1}^{L} \prod\limits_{j=1}^{D^l}p(\bbf_j^l|\bu_j^l, F^{l-1}, Z^l)q(\bu_j^l)$, where  $q(\bu^l_j) = \bN(\bu^l_j; {\bf m}^l_j,S^l_j )$ ~\cite{titsias2009variational,damianou13,GPbigdata}. Let ${\bf m}^l$ be a vector formed by concatenating the vectors ${\bf m}^l_j$ and $S^l$ be the block diagonal covariance matrix formed from $S^l_j$.      
We can formulate the \textbf{ELBO} by extending the methodology described in Section \ref{sect:gp} to multiple layers~\cite{damianou13,dsvi} as follows:
\begin{eqnarray}
\label{dsvi:eqn1}
& & L( \{{\bf m}^l , S^l\}_{l=1}^L) = \sum\limits_{n=1}^{N} \bE_{q(F_n^L)}[\log p(y_n|F_n^L) ]_ - \sum\limits_{l=1}^L KL[q(U^l) || p(U^l) ]  
\end{eqnarray}
where, the marginal distribution of the functions values for layer $L$ over all the data points is obtained as
\begin{eqnarray}
\label{dsvi:eqn2}
& & \hspace{-5mm}  q(F^L | \{Z^l,{\bf m}^l , S^l\}_{l=1}^L)=\int\limits_{F^1,F^2, \cdots F^{L-1}} \prod\limits_{l=1}^{L} q(F^l|F^{l-1},Z^l, {\bf m}^l , S^l)  dF^{1}\ldots dF^{L-1}
\end{eqnarray}
\noindent and the conditional distribution in (\ref{dsvi:eqn2}) is computed  as 
\begin{eqnarray}
\label{eqn:dsgp_conditional}
& & \hspace{-5mm} q(F^l|F^{l-1}, Z^l, {\bf m}^l , S^l) = \prod_{j=1}^{D^l} \int\limits_{\bu^l_j }p(\bbf_j^l|\bu_j^l,F^{l-1}, Z^l) q(\bu_j^l) d\bu_j^l = \prod_{j=1}^{D^l}  \bN(\bbf_j^l; {\bf \tilde{m}}^l_j, \tilde{V}^l_j)  \\
& & \hspace{-5mm} \text{ where } {\bf \tilde{m}}^l_j = \textit{m}^l_j(F^{l-1}) + K^l_{F^{l-1}Z^L}(K^l_{Z^lZ^l})^{-l}(\textbf{m}^l_j - \textit{m}^l_j(Z^l)) \text{ and } \\
& & \hspace{-5mm} \tilde{V}^l_j = K^l_{F^{l-1}F^{l-1}} - K^l_{F^{l-1}Z^l}(K^l_{Z^l Z^l})^{-l}(K^l_{Z^l Z^l} - S^l_j)(K^l_{Z^l Z^l})^{-l}(K^l_{F^{l-1}Z^l})^\top .
\end{eqnarray}

\noindent The marginal distribution in  (\ref{dsvi:eqn2}) is intractable, due to presence of stochastic term $\{F^{l-1}\}_{l=2}^{L}$ inside the conditional distributions $\{q(F^l|F^{l-1}, Z^l,{\bf m}^l , S^l)\}_{l=2}^{L-1}$ in a non-linear manner. This intractability results in the expected log likelihood in (\ref{dsvi:eqn1}) to be intractable even for Gaussian likelihood. We approximate it via Monte Carlo sampling as done in ~\cite{dsvi}.

As has been shown in \cite{dsvi}, the marginal variational posterior over function values in the final layer for $n^{th}$ data point, i.e 
$q(F_n^L)$ depends only on the $n^{th}$ marginals of all the previous layers. Each $F_n^l$ is sampled from $q(F_n^l|F_n^{l-1}, Z^l, {\bf m}^l , S^l)$ $= \bN(F_n^l; {\bf \tilde m}^l[n],\tilde{V}^l[n] )$, where ${\bf \tilde m}^l[n]$ ($D^l$ dimensional vector) and $\tilde{V}^l[n]$ ($D^l \times D^l$ diagonal matrix) are respectively the mean  and covariance of the $n^{th}$ data point over representations in layer $l$ and depends on $F^{l-1}_n$. Applying the ``reparametarization trick'' the sampling can be written as:
\begin{equation*}
F^l_n =  {\bf \tilde m}^l[n] + {\bm{\epsilon}}^l\odot{\tilde{V}^l[n]}^{\frac{1}{2}} ; \quad {\bm{\epsilon}}^l \sim \bN(\bm{\epsilon}^l;0,\mathbb{I}_{D^l}).
\end{equation*}
The lower bound  can be written as sum over data points and the parameters can be updated based gradients computed on a mini-batch of data. This enables one to use stochastic gradient  techniques for maximizing the variational lower bound. This stochasticity in gradient computation combined with the stochasticity introduced by the Monte Carlo sampling in variational lower bound computation results in the doubly stochastic variational inference method for deep GPs. 

\section{Convolutional Deep Gaussian Processes}

We combine the convolutional GP kernels~\cite{convgp} with deep Gaussian processes in order to obtain the convolutional deep Gaussian process (CDGP). A CDGP can capture salient features which are invariant to variations in the image through the convolutional structures and is simultaneously performing strong function learning through out its depth, all within a Bayesian framework. This results in a powerful well-calibrated model for tasks like image classification.

\subsection{Convolutional kernels}
Our starting point is the recently introduced convolutional Gaussian processes (CGP)~\cite{convgp}  where the function evaluation on an image is considered as sum of functions  over the patches  of the input image. Assuming there are  $P$  patches in $\bx$ with each patch $\bx^{[p]}$ to be $w\times h$ dimensional, CGP considers $f(\bx) = \sum_{p=1}^P g(\bx^{[p]})$. 
Placing a zero mean  ${\mathcal{GP}}$ prior over the function $g(\bx^{[p]})$, $g(\bx^{[p]})\sim {\mathcal{GP}}(0, k_g(\bx_i^{[p]},\bx_j^{[p]}))$, induces a zero mean ${\mathcal{GP}}$ prior over the function $f(\bx)$ with a convolutional kernel (Conv kernel) $k_f$,
\begin{eqnarray}
f(\bx) \sim {\mathcal{GP}}(0,k_{f}(\bx_i,\bx_j)), \qquad 
k_{f}(\bx_i,\bx_j) = \sum\limits_{p=1}^P\sum\limits_{p'=1}^P k_g(\bx_i^{[p]},\bx_j^{[p']}) . 
\end{eqnarray}
We refer to $k_g$ as the base kernel. Considering a convolutional kernel in computing the similarities between the images is useful in capturing non-local similarities among the images. The convolutional kernel  compares  one region in the image $\bx_i$ with another region in the image $\bx_j$, and could provide a high similarity even under transformations in the image. The kernel computation over patches $(\bx_i^{[p]},\bx_j^{[p']})$ considers similarity in a spatial neighborhood, whereas with other kernels (such as RBF kernel) only global similarity across images can be computed and fails to capture similarity in images due to transformations.        

Convolutional Neural Networks(CNNs) convolve image with multiple kernels (filters), apply a non-linear operation and then feature pooling (average, max) multiple times to learn discriminative features useful for the object detection task. Similar to CNN, the function $f(\bx)$  could be seen to perform average pooling of the non-linear feature maps produced by the patch response functions $g(\bx^{[p]})$.  This pooling operation results in convolution operation in kernel space. The convolutional kernel computation between two images $\bx_i$ and $\bx_j$ is expanded as 
\begin{equation}
k_f(\bx_i,\bx_j) =\sum\limits_{p'=1}^Pk_g(\bx_i^{[1]},\bx_j^{[p']}) + \ldots + \sum\limits_{p'=1}^P k_g(\bx_i^{[p]},\bx_j^{[p']}) + \ldots + \sum\limits_{p'=1}^Pk_g(\bx_i^{[P]},\bx_j^{[p']}) \nonumber .
\end{equation}
The convolution operation between $p^{th}$ patch of image $\bx_i$ (which now acts as a filter) and the image $\bx_j$ results in a convolution signal, where signal value at any point $p'$ is obtained by computing the dot product between the filter $\bx_i^{[p]}$ and patch $\bx_j^{[p']}$. This dot product is performed by the base kernel which transforms these  patches into feature vectors in a high dimensional space and computes the dot product between them in that space. 
Any $p^{th}$ summand is the sum of the convolution signal values obtained at all the points.

\subsection{Deep Gaussian processes with convolutional kernels}

Convolutional DGP considers multiple functions from a GP prior with convolutional kernels to form a representation of the image in the first layer. The function corresponding to $o^{th}$ representation for layer $1$ is obtained as
\begin{eqnarray}
& & f^1_o(\bx) = \sum_{p=1}^P g^1_o(\bx^{[p]}) \quad ; \quad g^1_o(\bx^{[p]}) \sim \mathcal{GP}(m_o^1(\bx^{[p]}), k^1_g(\bx_i^{[p]}, \bx_j^{[p]})) \\
& & f^1_o(\bx) \sim \mathcal{GP}(m_o^1(\bx), k^1_f(\bx_i, \bx_j)) \quad ; \quad k^1_{f}(\bx_i,\bx_j) = \sum\limits_{p=1}^P\sum\limits_{p'=1}^P k^1_g(\bx_i^{[p]},\bx_j^{[p']}). \ \ \  
\end{eqnarray}
Each output in layer $1$ captures different features of the image. The feature representations of the image obtained in the first layer are then mapped using a GP with convolutional or RBF kernel to obtain further representations. In general, the function corresponding to $o^{th}$ representation for layer $l$ is considered as
\begin{eqnarray}
\label{dcgp:convkernel}
& & f^l_o(F^{l-1}) \sim \mathcal{GP}(m_o^l(F^{l-1}), k^l_f(F^{l-1}_i, F^{l-1}_j)) \nonumber \\
& & k^l_{f}(F^{l-1}_i,F^{l-1}_j) = \sum\limits_{p=1}^P\sum\limits_{p'=1}^P k^l_g({F^{l-1}_i}^{[p]},{F^{l-1}_j}^{[p']}) .
\end{eqnarray}
The kernel matrices involved in the computation of the conditional distribution in eq. (\ref{eqn:dsgp_conditional}) such as $K^l_{F^{l-1}F^{l-1}}$, $K^l_{F^{l-1}Z^l}$ and ${K^l_{Z^l Z^l}}$  use the convolutional kernel defined in (\ref{dcgp:convkernel}). As before, $Z^l$ represents the inducing points associated with layer $l$ and has the same dimension as $F^{l-1}$. The variational lower bound expression and ``reparameterization trick'' remains the same as has been derived for deep GPs in Section \ref{sect:dgp}.
  
We also consider variants of the convolutional kernel such as weighted convolutional kernels (Wconv kernels)~\cite{convgp}.  It associates a weight with each patch which allows the kernel to provide differential weightage to the patches which is useful for object detection. The function  $f(\bx)$  in general for any layer is considered as 
\begin{eqnarray}
& & f(\bx)  = \sum_{p=1}^P w_p g(\bx^{[p]}) \quad ; \quad  k_{f}(\bx_i,\bx_j) = \sum\limits_{p=1}^P\sum\limits_{p'=1}^P w_p w_{p'} k_g(\bx_i^{[p]},\bx_j^{[p']}) .
\end{eqnarray}

\subsection{Reducing computational complexity through patch subsets} 
Convolutional kernels provide an effective way to capture the similarity across images, but are computationally expensive. Computing the similarity between two images involves $\mathcal{O}(P^2)$ computational cost, where $P$ is the number of patches in the input image or the feature representation. For the input image of size $W \times H$, it is of the order of $\mathcal{O}(WH)$ when stride length and patch sizes are small. This is costly even for image data sets such as MNIST and rectangles which contain images of size ($28 \times 28$). This makes the computations impractical on higher dimensional data such as Caltech101 ($250 \times 250$). This can be addressed to some extent using the idea of  treating the inducing points in the patch space~\cite{convgp}, where $Z^l_j\in \bR^{w \times h}$ rather than in the input space $\bR^{W \times H}$. In this case, computation of the entries in the matrix $K^l_{F^{l-1}Z^l}$ can be performed in $\mathcal{O}(P)$ time, and that of $K^l_{Z^l Z^l}$ can be performed in constant time. 
\begin{eqnarray}
 & & K^l_{F^{l-1}Z^l}[i,j] = k^l_f(F^{l-1}_i, Z^l_j) = \sum_{p=1}^P k_g^l({F^{l-1}_i}^{[p]}, Z^l_j) \\
 & & {K^l_{Z^l Z^l}}[i,j] = k_g^l(Z^l_i, Z^l_j)
\end{eqnarray}
However, computation of the entries in the matrix $K^l_{F^{l-1}F^{l-1}}$ matrix which appears in the conditional distribution in (\ref{eqn:dsgp_conditional})  still requires $\mathcal{O}(P^2)$ computations for the first layer making it a costly operation. This makes the approach practically inapplicable to high dimensional data sets such as Caltech101 even with a reduced image size. Moreover in these images, a lot of information will be shared by overlapping patches and will be redundant for the computation of the similarity across images. We propose to use random sub-sampling of the patches in computing the convolutional kernel for the entries in the matrix  $K^l_{F^{l-1}F^{l-1}}$ and $K^l_{F^{l-1}Z^l}$. Let ${S,S'} \subset \{1,2, \ldots, P\}$ represent the random subsets. For the $o^{th}$ representation of layer $1$ ($F^{0} = X$), we consider the covariance functions to be as follows

\begin{eqnarray}
& & f^1_o(\bx) = \sum_{p \in S} g^1_o(\bx^{[p]}) \\ 
& &  k^1_{f}(\bx_i,\bx_j) = \sum\limits_{p \in S}\sum\limits_{p' \in S'} k^1_g(\bx_i^{[p]},\bx_j^{[p']}) \text{ and }
\end{eqnarray}
\begin{eqnarray}
& & k^1_f(\bx_i, Z^1_j) = \bE_g[f^1_o(\bx_i)g^1_o(Z^1_j)] = \bE_g [\sum_{p \in S} g^1_o(\bx^{[p]})g^1_o(Z^1_j)]\\
&&\hspace{+17mm}= \sum_{p \in S} k^1_g(\bx^{[p]},Z^1_j)
\end{eqnarray}
This reduces the cost of computing the matrix $K^l_{F^{l-1}F^{l-1}}$ for layer $1$ to $\mathcal{O}(|S||S'|)$ where the size of the subsets $|S|,|S'| \ll P$. Computational speedup achieved through random sub-sampling of patches is testified in our experiments on Caltech101.

\section{Experiments}

We evaluate the generalization performance of  the proposed model, convolutional deep Gaussian processes (CDGP), on various image classification data sets, namely MNIST, Rectangle-Images, CIFAR10 and Caltech101. We consider different kernel architectures of the proposed CDGP model and compare it with sparse GPs (SGP) \footnote{Results as reported in \cite{dsvi}.}, deep GP (DGP) models with RBF kernel and with convolutional GPs (CGP) with different convolutional kernels.  The convolutional deep GP uses the same inference procedure as in deep GP (``re-parameterization trick'') and uses an a priori fixed inducing input points by considering centroids of the clustered images \cite{dsvi}. The inducing points and the linear mean function for each of the inner layers is obtained using the singular value decomposition approach  mentioned in \cite{dsvi}. The number of inducing points is taken to be 100. We follow the same approach for convolutional GPs also to maintain a fair playground. The kernel parameters are kept the same across various outputs in a layer while it is different across the layers. The number of outputs in the latent layers is taken to be $30$ for MNIST and $50$ for other datasets (except for the final layer which will be equal to the number of classes). For the models considering convolutional kernels, the patch size is taken to be $3 \times 3$ with a stride length of $1$ for the rectangles data while a patch size of $5 \times 5$ is considered for the rest of the data sets. We consider the RBF kernel as the base kernel $k_g$ for all our experiments. The approaches are compared in terms of their accuracy in making predictions on the test data and the negative log predictive probability (NLPP) on test data which considers uncertainty in predictions. 
The code has been developed on top of GPflow~\cite{GPflow2017} framework with ADAM~\cite{AdamAM} optimizer to learn the kernel and variational parameters by maximizing the variational lower bound.  The variational mean parameters are initialized to $0$, variance parameters to $1e^{-5}$ and length-scales are initialized to $2$ for MNIST and $10$ for other datasets.

\begin{table}[t]
\caption{Comparison of SGP, DGP, CGP and CDGP approaches with different architectures on the MNIST data set along with the kernels used by GP in each layer.  }
\label{tab:mnist}
\begin{center}
    \begin{tabular}{ | p{2cm} | p{1.5cm} | p{1.5cm} | p{1.5cm} |  p{1.5cm} | p{2cm}|p{1.5cm} |} 
    \hline
    Model & Layer 1 & Layer 2 & Layer 3 & Layer 4 & Accuracy\% & NLPP \\ 
    \hline 
    SGP  & RBF  & -- & -- &  -- & 97.48 & -- \\ 
    DGP1 & RBF & RBF& -- &  -- &97.94 & 0.073 \\ 
    DGP2 & RBF  & RBF& RBF & -- & 97.99 & 0.070\\        

CGP1 &   Conv  & -- & -- & -- & 95.59 & 0.170 \\
	CGP2 & Wconv &  -- & -- & -- & 97.54 & 0.103\\ 
	\bf{CDGP1} &  Wconv & RBF &  --&--&   \bf{98.66} & \bf{0.046} \\ 
     {CDGP2} &  Conv& RBF &  --&--&  {98.53} & 0.536 \\
     {CDGP3} &  Conv& RBF &  RBF &--&   {98.40} & 0.055 \\
     {CDGP4} &  Conv& RBF &  RBF & RBF &   {98.41} & 0.051 \\
     {CDGP5} &  Wconv& Wconv &  RBF &--&   {98.44} & 0.048 \\
     {CDGP6} &  Wconv& Wconv &  RBF & RBF &   {98.60} & 0.046 \\ \hline
   \end{tabular}
\end{center}  
\end{table}
\begin{table}[t]
\caption{Comparison of SGP, DGP, CGP and CDGP approaches with different architectures on the Rectangles-Image data set along with the kernels used in by GP in each layer.  }
\label{tab:rectangle}
\begin{center}
    \begin{tabular}{ | p{1.5cm} | p{1.5cm} | p{1.5cm} |  p{1.5cm} | p{2cm}|p{1.5cm}|} 
    \hline
    Model & Layer 1 & Layer 2 & Layer 3  & Accuracy\% & NLPP \\ 
    \hline 
    SGP &   RBF  & -- & --  & 76.1 & 0.493 \\    
	 {DGP1} &  RBF & RBF &  --&  76.93 & 0.478 \\ 
     {DGP2} &  RBF & RBF &  RBF &  76.98 & 0.476 \\
      CGP & Wconv &  -- & -- &  71.06 & 0.602 \\ 
     \bf{CDGP1} &  Wconv & RBF &  -- &   \bf{79.74} & \bf{0.422} \\
     {CDGP2} &  Wconv & RBF &  RBF &  77.95 & 0.449  \\
     \hline
   \end{tabular}
\end{center}
\end{table}

\subsection{MNIST-10}
We performed experiments with MNIST  dataset with 10 classes corresponding to the digits $0-9$.  We consider the standard train/test split with 60K training and 10K test images. We considered CDGP and DGP models with various architectures as described in Table~\ref{tab:mnist}. Parameters of the model are learned by running the ADAM optimizer for $400$ epochs with $0.01$ step size and a mini-batch of 1000. Experimental comparison indicates that the proposed CDGP models with 2 layers, first layer with a weighted convolutional kernel and the second layer with an RBF kernel gave the best performance, an accuracy of $98.66$ and an NLPP score of $0.0463$. Second best performance was given again by a CDGP model with 4 layers, 2 weighted convolutional kernels followed by 2 RBF layers. We could observe that all the CDGP models performed better than the DGP and CGP models in the MNIST data.  We also conducted experiments with the combinations of two RBF kernels  with length scales initialized to 2 different values $0.01$ and $10$, as the base kernel in a convolutional kernel. The approach gave an accuracy of $98.46$. We found that the learned length scales are also quite far apart which shows that one RBF kernel is trying to capture long distance correlations  while the other one captures short distant correlations. This  did not result in better results as MNIST is quite simple dataset for which capturing such information might not be necessary.

\subsection{Rectangles-Image}

We consider the rectangles-image data set used in \cite{dsvi}, where a rectangle  of varying height and width is placed inside images. The patches in the border and inside of the rectangle and the background patches are sampled to make the rectangle hard to detect~\footnote{Rectangles-image data is different from the simpler rectangles data used in \cite{convgp}, where a random size rectangle is placed in black background with the pixels corresponding to the border of the rectangle in white, while that of inside in black. }.  The task is to classify if a rectangle in an image has a larger height or width. The data set consists of 12K training images and 50K test images, and is known to require deep architectures for correct classification. We consider two different architectures of CDGP, and compare it against sparse GPs, deep GPs with 2 and 3 layers and convolutional GPs. Parameters of the models are learnt by running the ADAM optimizer for $200$ epochs with $0.01$ step size and a mini-batch of 1000. Experimental comparison across different approaches is provided in Table~\ref{tab:rectangle}. We could observe that the proposed CDGP model with 2 layers, first layer using a weighted convolutional kernel and the second layer using an RBF, provided the best performance beating DGP, CGP and SGP models by a large margin. To the best of our knowledge, this is the highest accuracy reported by a GP model on the rectangles-image data. 
This indicates the usefulness of the representation learning capability of CDGP  model for complex image classification.

\begin{table}[t]
\caption{Comparison of SGP, DGP, CGP and CDGP approaches with different architectures on the CIFAR10 data set along with the kernels used by GP in each layer.  }
\label{tab:cifar10}
\begin{center}
    \begin{tabular}{ | p{1.5cm} | p{1.5cm} | p{1.5cm} |  p{1.5cm} | p{2cm}|p{1.5cm}|} 
    \hline
    Model & Layer 1 & Layer 2 & Layer 3  & Accuracy\% & NLPP \\ 
    \hline 
	 {DGP1} &  RBF & RBF &  --&  42.20 & 3.2579 \\ 
     {DGP2} &  RBF & RBF &  RBF &  40.13 & 3.5785 \\
      \bf{CGP} & Wconv &  -- & -- &  \bf{55} & -- \\ 
     {CDGP1} &  Wconv & RBF &  -- &   51.74 & 2.4893 \\
     {CDGP2} &  Wconv & RBF &  RBF &  51.59 & 2.4607 \\
     \hline
   \end{tabular}
\end{center}
\end{table}

\subsection{CIFAR-10}
The CIFAR-10 dataset~\cite{cifar10} consists of total 60K images out of which 50K are used as training images while the rest 10K images are being used for testing. The dataset contains colored images of objects like airplane, automobile, etc. There are 10 classes in total having 6K images per class. The dimensionality of each image is $32\times32\times3$ (3 is for channels). We compare the performance of CDGP, DGP and CGP models in Table~\ref{tab:cifar10}. Parameters of the models are learned by running the ADAM optimizer for $200$ epochs with  a mini-batch size of 40\footnote{Learning took around 11 hours on Nvidia GTX 1080 Ti GPU, while the best results reported in ~\cite{convgp} is obtained after running the optimization for 40 hours.}.

We observe that DGP models gave a relatively low performance on the CIFAR10 datasets. Equipping DGP models with convolutional kernels have boosted the performance by $10\%$ showing the effectiveness of convolutional kernels for image classification. However, CDGP models were not able to obtain a performance close to CGP. This could be an indication that, for this particular dataset, the properties of a single-layer CDGP i.e, CGP is enough to learn a good classifier.  
In fact, the previous experiments have shown that 2-layer CDGPs typically result in the best accuracy (in comparison with deeper models), implying that a CGP has already very large capacity for classification and therefore the addition of one layer is usually enough to improve on the results.

\begin{table}[t]
\caption{Comparison of Training time required for different CDGP architectures with different number of patches on the Caltech-101 dataset.}
\label{tab:caltech101}
\begin{center}
    \begin{tabular}{ | p{2.7cm} | p{1.3cm} | p{1.3cm} |  p{1.3cm} | p{2cm}|p{1.5cm}|p{1.5cm}|} 
    \hline
    Model & Layer 1 & Layer 2 & Layer 3  & Training time & Accuracy\% & NLPP \\ 
    \hline 
     {CDGP1(All patches)} &  Wconv & RBF & -- & 11 hrs 18 min  &20.39 &6.5811   \\
     {CDGP2(All patches)} &  Wconv & RBF &  RBF & 12 hrs 2min    &19.51  &6.787 \\
     {CDGP1(Random patches)} &  Wconv & RBF &  -- & 1 hr 15 min  &  20 & 6.7009   \\
     {CDGP2(Random patches)} &  Wconv & RBF &  RBF & 1hr 19 min & 18.82 & 7.0473 \\
	\hline
   \end{tabular}
\end{center}
\end{table}

\subsection{Experiments with Random Sub-sampling of Patches on Caltech-101 Dataset}
Computation of convolutional kernels becomes prohibitive on data sets such as Caltech101~\cite{caltech} with very high dimensionality. 
It consists of 101 classes with 20 images per class for training and 10 images per class for testing. The size varies slightly for each image in the actual dataset but is roughly around  300 $\times$ 200 pixels per channel. The images are colored so each image has 3 channels. The experiments are conducted on images resized to $50\times50\times3$. Instead of taking all the patches of the image for computing the convolutional kernel, we randomly picked up one-tenth of the total number of image patches for computing the kernel. This resulted in a very significant speed-up in learning time without much loss in accuracy, as can be seen from table \ref{tab:caltech101}~\footnote{We ran the experiments on Nvidia GTX 1080 Ti GPU.}. The test accuracy obtained with CDGP1 is $20.39\%$ and time taken for training is 11 hrs 18min. On the other hand, for the same model considering random subset of image patches,  training time drops to only 1 hr 15 min with an accuracy drop of only $0.39\%$ making it around 10 times faster. Similar phenomenon has been observed in case of CDGP2 considering random subset of patches, where training time improved  from 12 hrs 2 min to 1 hrs 19 min with an accuracy drop of just $0.69\%$, providing a speedup of around 10 hours. The classification accuracies of the models presented in Table~\ref{tab:caltech101} are low due to resizing of the original image to size $50 \times 50$, resulting in loss of information. As a future work, we will conduct experiments by keeping the  original image size, and study the effectiveness of random sub-sampling of patches and generalization performance of the proposed approach.

\section{Conclusion}
Deep GP models provide a lot of advantages in terms of capacity control and predictive uncertainty, but they are less effective in computer vision tasks. Commonly used RBF kernels in the DGP models fail to capture  variations in  image data and are not invariant to translations. In this paper we proposed a DGP model which captures convolutional structure in image data using convolutional kernels. Our model extends the convolutional GPs with the ability to learn hierarchical latent representations making it a useful model for image classification. We incorporated different types of convolutional kernels \cite{convgp} in the DGP models and demonstrated their usefulness for image classification in benchmark data sets such as MNIST, Rectangles-Image and CIFAR10. In the future, we plan to develop methods to further reduce the cost of convolutional kernel computation and memory requirements of the CDGP model for high dimensional datasets. This will allow us to consider a higher mini-batch size, leading to reduced stochastic gradient variance and faster convergence of the optimization routine. We found that increasing the number of layers in CDGP did not bring much improvements in performance contrary to what we expected. We hope that our future research on faster and more effective variational inference techniques will address these limitations with convolutional DGPs.

\newpage
\bibliographystyle{unsrt}
\bibliography{condgp}

\end{document}